%% file: CLICKBAIT.tex
\documentclass[a4paper,11pt,twocolumn,twoside]{article}
\usepackage[dvips]{graphicx}
\usepackage{sepln_en}
\usepackage{fullname}
\usepackage{pdfpages}
\usepackage{adjustbox}
\usepackage{booktabs}
\usepackage{caption}
\usepackage{url} 
\usepackage{float}
\usepackage{hyperref}
\usepackage{multirow}
\usepackage{listings}
\usepackage{algpseudocode}
\usepackage[english]{babel}
\usepackage{gb4e}\noautomath
\usepackage{placeins}
\input epsf

\title{\raisebox{-.25\height}{\includegraphics[width=0.7cm]{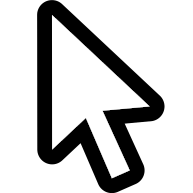}}NoticIA: A Clickbait Article Summarization Dataset in Spanish}

\author {\textbf{Iker García-Ferrero}, \textbf{Begoña Altuna}\\
HiTZ Basque Center for Language Technology - Ixa \\ University of the Basque Country UPV/EHU \\
{\{iker.garciaf,begona.altuna\}@ehu.eus}
}

\seplnabstract{We present NoticIA, a dataset consisting of 850 Spanish news articles featuring prominent clickbait headlines, each paired with high-quality, single-sentence generative summarizations written by humans. This task demands advanced text understanding and summarization abilities, challenging the models' capacity to infer and connect diverse pieces of information to meet the user's informational needs generated by the clickbait headline. We evaluate the Spanish text comprehension capabilities of a wide range of state-of-the-art large language models. Additionally, we use the dataset to train ClickbaitFighter, a task-specific model that achieves near-human performance in this task.}

\seplnkey{Clickbait, Summarization, Dataset, Spanish.}

\firstpageno{1}

\begin{document}


\setlength\titlebox{9.5cm} 

\maketitle

%

\section{Introduction}

In the digital age, the expansion of clickbait headlines has become a pressing concern for both media consumers and publishers. Clickbait refers to sensationalized or misleading headlines designed to lure readers into clicking on a link, often at the expense of accurate reporting and journalistic integrity. In clickbait headlines the actual content is exaggerated and distorted, driving readers to misinformation and confusion. Those headlines create a need in users to discover specific information, compelling them to click on the article, although the low quality of the piece of news may make them quickly exit the website without engaging with other ads or articles. 

The clickbait headlines often lead readers to enter a webpage and scroll down it while they try to find the promised information they are searching for. Such articles often contain a plethora of irrelevant details, with the main idea typically buried at the end. The ultimate goal of clickbait-led news is to show as many advertisements as possible to the readers in a way of trying to increase the advertisement revenue of the site. This practice tends to be annoying for the readers and undermines their trust in online news outlets. For publishers, while clickbait may offer short-term gains in terms of increased web traffic and potential advertising revenue, it carries the risk of damaging their reputation and alienating their reader base. It also negatively impacts advertising revenue for legitimate content creators, who might see their web traffic reduced.

\begin{figure*}[t]
\centering
\includegraphics[width=0.95\linewidth]{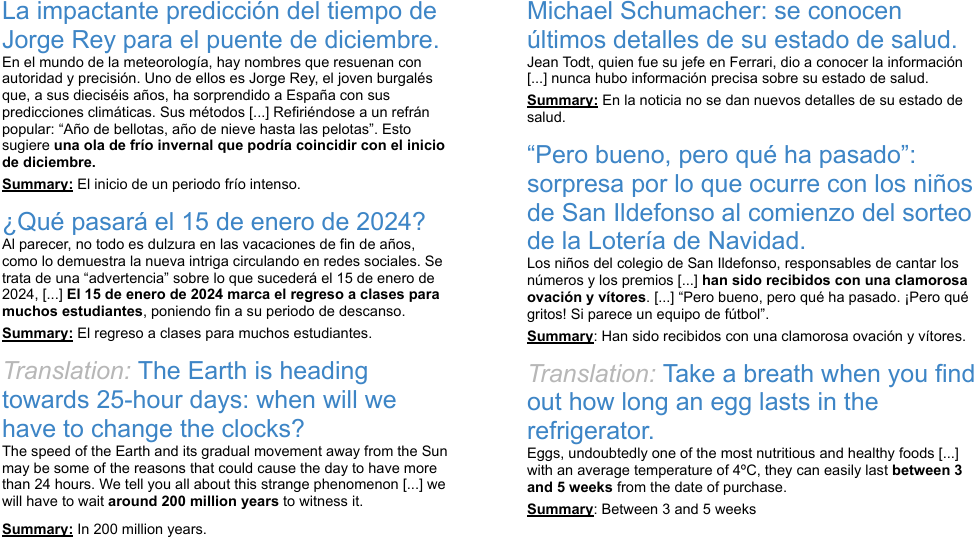}
\caption{Examples of clickbait headlines from NoticIA. The headline is followed by a long article in which the answer to the headline is located at the end of the article. We translated two examples into English for illustration.}
\label{fig:Clickbait_Examples}
\end{figure*}

%
%
%

The task of summarizing these low-quality articles with clickbait headlines constitutes a great benchmark for Large Language Models (LLMs). Unlike general domain summarization, this task first requires the model to understand the headline and the information it promises. The summary depends on accurately interpreting the headline, as the failure to do so will lead to an incorrect summary. Then, it must successfully navigate through the filler content to identify and extract the key information or idea that is often hidden, not immediately obvious, or, in some cases, completely missing because the headline was misleading. This task demands advanced text understanding and summarization abilities and challenges the models' capacity to infer and connect diverse pieces of information to fulfil the user's informational needs as suggested by the clickbait headline. This challenge can be described as an ``ultrasummary'' generation task, where articles, designed to be as lengthy as possible, are summarized into a single sentence or even a single word. Figure \ref{fig:Clickbait_Examples} illustrates examples of clickbait headlines from our dataset, together with the human-written summaries. We also provide two translated examples to make the task description more accessible.


To the best of our knowledge, \raisebox{-.25\height}{\includegraphics[width=0.5cm]{logo.png}}NoticIA is the only clickbait dataset that provides headline-news body-ultrasummary triplets that can be used to assess the performance of information retrieval systems that will help extract crucial information.  NoticIA comprises 850 Spanish news articles featuring single-sentence summaries written by humans. Our contributions are the following:
\begin{itemize}
    \item We introduce a dataset containing 850 Spanish news articles, each with a clickbait headline, news body, and human-written summary.\footnote{\url{https://hf.co/datasets/Iker/NoticIA}}\footnote{\url{https://hf.co/datasets/somosnlp/NoticIA-it}}
    \item We evaluate several state-of-the-art text-to-text Large Language Models (LLMs) in zero-shot settings.\footnote{\url{https://github.com/ikergarcia1996/NoticIA}} This evaluation demonstrates the text-understanding capabilities of these models in Spanish, addressing the current scarcity of high-quality text understanding benchmarks and evaluations for LLMs in this language.
    \item We fine-tune and publicly release LLMs trained on our dataset.\footnote{\url{https://hf.co/Iker/ClickbaitFighter-2B}}\footnote{\url{https://hf.co/Iker/ClickbaitFighter-7B}}\footnote{\url{https://hf.co/Iker/ClickbaitFighter-10B}}  This model family exhibits high proficiency in summarizing clickbait articles and has the potential to empower individuals by providing them with tools for critical thinking and discernment amidst the vast amounts of online content. By making these models public, we aim to exert pressure against the use of deceptive tactics by online news providers to increase advertising revenue.
\end{itemize}





The coming sections of the paper are structured as follows. In section \ref{sec:related} we go through the most relevant works on the topic. In section \ref{sec:dataset} the corpus building effort is described. The experimental set-up is presented in section \ref{sec:setup} and its development is described in section \ref{sec:experiments}. Finally, the final remarks are done in section \ref{sec:conclusion}.

\section{Related works}\label{sec:related}

Most of the efforts aimed at combating clickbaits have focused on clickbait headline detection. Although it is over a decade that the task has attracted the interest of scholars, the Clickbait Challenge \cite{potthast2018} in 2017 was a milestone in the advancement of the field. There were 12 teams taking part and the Webis Clickbait Corpus 2017 was released, which contained 38,517 annotated tweets and that has been used since in a list of clickbait detection works such as the work by \namecite{ZHENG2021}. In addition, clickbait detection efforts have multiplied in the last lustrum as in the case of \namecite{pujahari2021}, \namecite{Liu2021} or \namecite{Wang2023}.


Our work, however, is more closely related to the clickbait spoiling task. The PAN Clickbait Challenge at SemEval 2023 \cite{froebe:2023d} consisted in classifying the types of spoilers according to their structure and in generating spoilers according to their type. 23 teams submitted their systems, which were tested on the Webis Clickbait Spoiling Corpus 2022, a 5,000 headline-news-spoiler corpus collected from social media.

The task can also be designed in a question-answering manner. In \namecite{Heiervang2022} a title-answering method is proposed. Fine-tuned LLMs are requested to extract an abstractive summary from the news body to answer the question in the clickbait headline. \namecite{johnson2022saved} consider both abstractive and extractive summaries to provide an answer for each headline. They fine-tune a RoBERTa model to perform the task. \namecite{Maharani2023}, instead, configure the task as a single or multi-span information extraction task for question answering.

Misleading headlines in Spanish news have also been a matter of study as in \namecite{Sepulveda2023}. The goal of the task is to identify whether a headline contains any contradiction in relation to the news body and classify the types of contradictions. For this, the ES\_Headline\_Contradiction dataset is built, which contains 18,542 news headline-body pairs. A series of BERT-based models are fine-tuned for both detection and classification and they achieve highly satisfactory results in both tasks.


\section{\includegraphics[width=0.3cm]{logo.png}NoticIA dataset}\label{sec:dataset}
The NoticIA dataset comprises 850 clickbait headline, news body and summary triplets. In this section, we will describe the construction of the dataset.

\subsection{Corpus collection}
Clickbait headlines have proliferated on the web in recent years. However, gathering clickbait headlines is an arduous task. To avoid the need for massive crawling of articles and the subsequent post-processing step to detect articles with clickbait headlines, we utilized two readily available sources of clickbait headlines to build the dataset. The first source is the posting on the X social media platform by the \href{https://twitter.com/ahorrandoclick1}{@ahorrandoclick1} user, who manually summarizes articles with clickbait headlines, from which we gathered 725 articles. Secondly, during the early development of this dataset, we fine-tuned a model to summarize clickbait articles. We used this model to develop a publicly available web app\footnote{\url{https://iker-clickbaitfighter.hf.space/}} 
that users could use to summarize articles given the article's URL. From the user inputs, we selected an additional set of 125 articles, for a total of 850 articles in the dataset.

\begin{figure}[t]
   \centering
   \includegraphics[width=\linewidth]{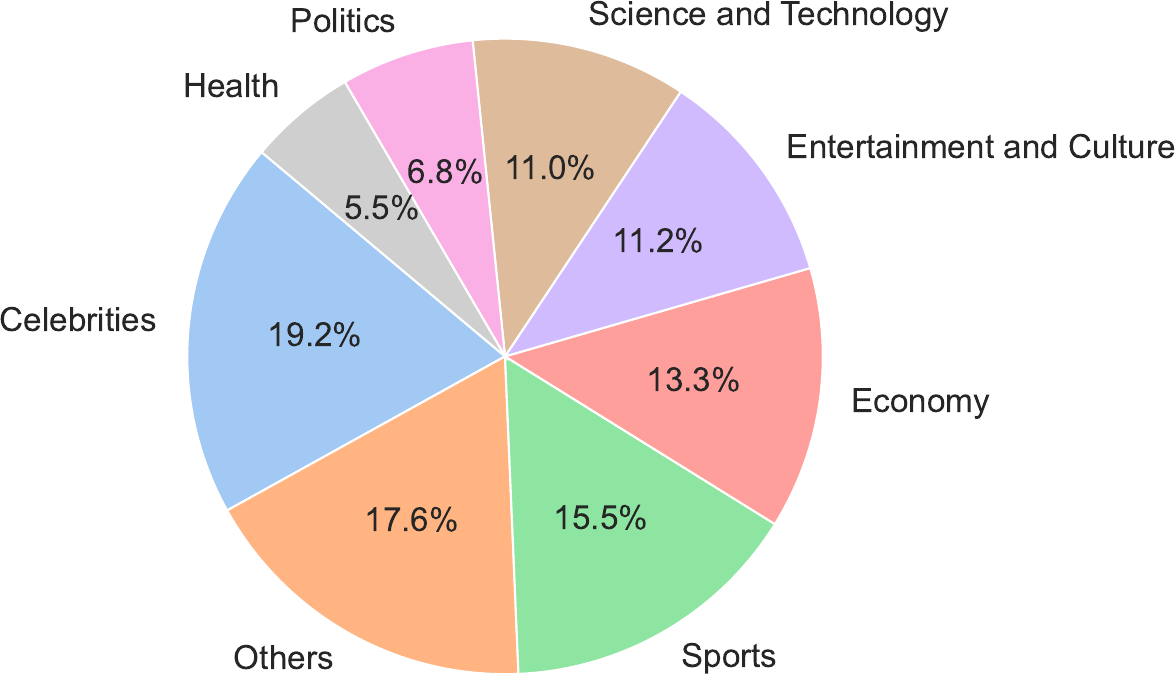}
   \caption{Category of the articles in the dataset.}
   \label{fig:NoticiaCategories}
 \end{figure}

The dataset comprises articles from a wide range of categories, as depicted in Figure \ref{fig:NoticiaCategories}. All articles are in Spanish, and approximately 83\% of the dataset originates from Spanish newspapers, while the remaining 17\% comes from Latin American newspapers. The detailed sources of the dataset can be found in Appendix \ref{apx:Sources}.

\subsection{Corpus annotation}

We employed human annotators to create gold standard summaries for the dataset. These annotators were PhD students with native-level competence in Spanish. They were instructed to read the headline and identify the question that the clickbait posed for the reader. Subsequently, they were tasked with finding the excerpt of the text that provided the answer to the clickbait headline. The annotators were asked to produce the shortest possible summary that responded solely to the clickbait headline, excluding any other information. Additionally, whenever possible, they were encouraged to directly quote the original text, especially when the summary included direct statements made by individuals. The full guidelines are available in Appendix \ref{apx:guidelines}. The annotation process took approximately 40 hours. Table \ref{tab:dataset_len} lists the average word counts for the headlines, article bodies, and summaries. On average, the article body was summarized with a 98\% reduction.

\input{Tablas/dataset_lens}


\subsection{Dataset quality assessment}
\label{sec:DataQuality}

\input{Tablas/dataset_validation}

The corpus annotations have been used to measure the quality of the distributed data and guidelines.\footnote{\url{https://hf.co/datasets/Iker/NoticIA_Human_Validation}} The test partition of our dataset was annotated by a second annotator to assess whether humans can provide aligned answers based on the guidelines. A summary of the inter-annotator agreement assessment is presented in Table \ref{tab:dataset_validation}. 

The overall agreement between the annotators has been high as they have provided the exact same answer in 26\% of the cases and have provided answers that partially share information in 48\% of the cases. This may imply that, although the news in our dataset tend to present the information intricately, it was easy for humans to find the information the headline refers to.
\input{Tablas/human_examples}
In the case of the partially overlapping answers, some may be considered to carry the same information, while others show complementary information as can be seen in the examples in Table \ref{tab:human_examples}.


We have also identified a list of cases in which the annotators have offered different but equally valid answers, which constitutes the 18\% of the cases. For example (\ref{ex1}) the annotators have chosen different text strings to refer to the same monetary prize. In (\ref{ex2}), instead, answers are not directly related as one refers to cold air and the other to unusually low temperatures.

\begin{exe}
    \ex \label{ex1} Headline: El 31938, tercer premio de la Lotería de Navidad 2023: esto es lo que cobrarás.
    \begin{xlist}
\exi {Annotator A:} 50.000€ por décimo.
\exi {Annotator B:} 500 mil euros por serie.
\end{xlist} 
\ex \label{ex2} Headline: Mario Picazo alerta de las consecuencias que tendrá el debilitamiento del vórtice polar en España.
    \begin{xlist}
\exi {Annotator A:} El aire muy frío confinado en el Ártico puede ser liberado hacia latitudes más bajas.
\exi {Annotator B:} Unas temperaturas que estarían muy por debajo de lo que sería habitual.
\end{xlist} 
\end{exe}

For what regards the guideline assessment, we venture to say that the guidelines were overall unambiguous and that the request of selecting the minimum amount of words to generate a valid summary will always be interpretable. As we could see from the annotators' answers, the minimum extent could be understood as the minimal well-formed sentence or the focus of the question in the headline  (\ref{ex3}).

\begin{exe}
\ex \label{ex3} Headline: Josep Pedrerol hace frente a la pérdida de un rostro conocido de `El Chiringuito de Jugones': ``Nunca olvidaré".
    \begin{xlist}
\exi {Annotator A:} Se ha ido el periodista Borja Mazarro.
\exi {Annotator B:} Borja Mazarro.
\end{xlist} 
\end{exe}

Finally, we identified 8 cases of disagreement. In 3 cases, one of the annotators produced an incorrect summary, likely due to fatigue after annotating multiple examples. In the remaining 5 cases, the disagreement stemmed from contradictory information in the article and different interpretations of this information (\ref{ex4}). In these instances, determining the correct summary is subject to the reader's interpretation. In any case, we consider the annotator agreement to be very high, and further experiments in Section \ref{sec:Consistency} demonstrate that there is a very high correlation on model performance when evaluated using the summaries produced by different annotators. 

\begin{exe}
\ex \label{ex4} Headline: El motivo por el que `Espejo Público' ha prescindido de un mítico colaborador: ``La gota que ha colmado vaso"
    \begin{xlist}
\exi {Annotator A:} Se refieren a Fran Rivera, el motivo fue aparecer en Telecinco.
\exi {Annotator B:} Rivera ya no aportaba mucho contenido al programa.
\end{xlist} 
\end{exe}

\section{Experimental Setup}\label{sec:setup}
In this section, we define the evaluation protocol we use to measure the performance of Large Language Models (LLMs) on our dataset.

\subsection{Models}
\input{Tablas/models_small}
We evaluate a diverse set of large language models (LLMs), ranging from 2 billion to 70 billion parameters. All the models assessed have been fine-tuned for instruction following. The models include  Deepseek \cite{deepseek}, LLama2 \cite{llama2}, Llama3 \cite{llama3modelcard}, Mistral \cite{mistral}, Solar \cite{solar}, Mixtral \cite{mixtral}, StableLM  \cite{stablelm-zephyr}, Yi \cite{ai2024yi} and Qwen \cite{qwen}, as well as further fine-tuned versions of these models which have improved capabilities. The models are listed in Table \ref{tab:model_sources_small}. This selection of models spans the most popular architectures at the time of writing and includes the current best-performing open-source models. Yet, it is important to note that most of these models were pretrained with a primary focus on the English or Chinese languages. Our comprehensive evaluation aims to shed light on the performance of the current state-of-the-art LLMs in Spanish.  In addition to these open-source models, we also evaluate the GPT-3.5, GPT-4 and GPT-4o commercial products. Although these models are not designed for research, and many details about them remain undisclosed, given their popularity and widespread usage,  we consider it valuable to assess their performance on our Spanish ultra-summarization task. We provide further details on the models we use in Appendix \ref{apx:ModelDetails}.

\subsection{Task Formulation}
\label{sec:TaskFormualation}
\begin{figure}[ht]
   \centering
   \includegraphics[width=\linewidth]{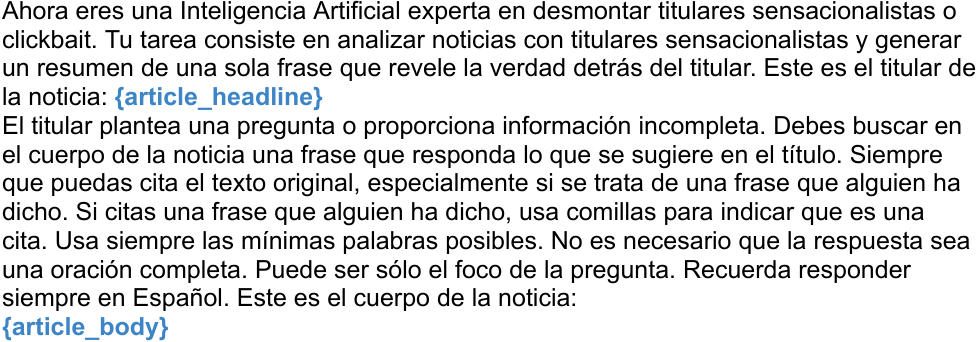}
   \caption{Input prompt used to generate summaries. The prompt defines the task and guidelines.}
   \label{fig:Prompt}
 \end{figure}

We adopt a zero-shot prompt evaluation setup, wherein we construct an instruction that describes the task and the annotation guidelines. The model receives the headline and the article body, and we expect a summary of the article as the output. The prompt used as input for the models is illustrated in Figure \ref{fig:Prompt}, a linguist expert on the summarization task crafted this prompt. The lengthy nature of the inputs, with article bodies averaging 550 words, precludes experimentation with more advanced prompting techniques, such as few-shot learning---this is, providing the model with a few summary examples---or Chain-of-Thought techniques. In the case of models that have been fine-tuned as dialogue systems, these utilize different tokens to represent a conversation, such as using markers like ``<bot>'' and ``<human>'', or custom system initial prompts. To accommodate these models, we format the input according to the recommendations provided by the authors. The input prompt is set to the user's input, and we expect the summary to be produced as the assistant's output. We do not specify any system-defined prompts and, for all the experiments, we use greedy search to generate text.

\begin{figure*}[ht]
   \centering
   \includegraphics[width=\linewidth]{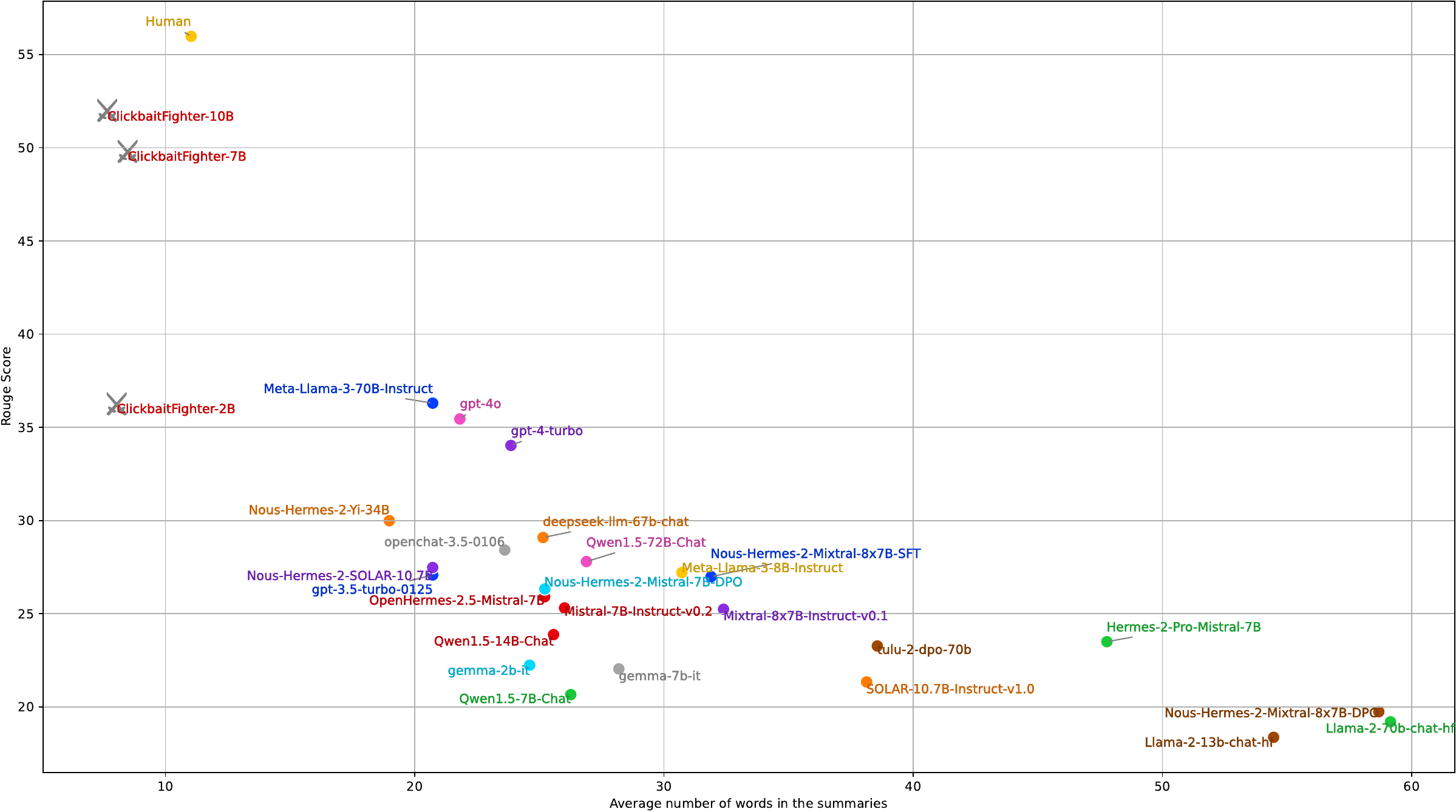}
   \caption{ROUGE score and average summary lengths for all models evaluated in our dataset. The Y-axis represents the ROUGE score, while the X-axis indicates the average number of words in the summaries. A higher ROUGE score and a shorter summary length are considered optimal.}
   \label{fig:Results}
 \end{figure*}

\subsection{Metrics}
\label{sec:Metrics}
As standard in summarization tasks, we use the ROUGE score metric \cite{lin-2004-ROUGE} to automatically evaluate the summaries produced by the models. ROUGE is a recall-oriented summarization metric that assesses the quality of summarization systems by determining how much of the basic units in the reference summaries\footnote{First annotator's summaries in this work.} appear in the machine-generated summaries. Our primary metric is ROUGE-1, which considers whole words as the basic units. To compute the ROUGE score, we lowercase both summaries and remove punctuation. 
In addition to the ROUGE score metric, we also consider the average length of the summaries. For our task, we aim for the summaries to be concise, which is an aspect the ROUGE score does not evaluate. Longer sentences may increase the chance of overlapping words, potentially affecting the ROUGE score. Therefore, we consider both ROUGE-1 and average summary length when evaluating the models. Our goal is to identify a model that achieves the \textbf{highest possible ROUGE score with the shortest summary length}, balancing quality and brevity.

Recent works \cite{min-etal-2023-factscore} have proposed advanced metrics for the automatic evaluation of general-domain summaries. However, our dataset significantly deviates from general domain summarization. Our summaries are highly concise, and we consider any information in the summary that does not contribute towards answering the clickbait headline as incorrect, even if this information appears in the article. Therefore, the set of possible correct summaries is very small, with any accurate summary necessarily including a specific set of words. Thus, we consider the ROUGE score as the best metric for evaluation, along with average sentence length to ensure that the model does not produce overly verbose summaries or include unnecessary information.

\section{Experiments} \label{sec:experiments}
In this section, we present the evaluation of a broad range of LLMs in both zero-shot and fine-tuning scenarios using the NoticIA dataset. Figure \ref{fig:Results} presents a summary of all the results, both in the zero-shot and fine-tuning (ClickbaitFighter models) settings. In this chart, the Y-axis represents the ROUGE score, while the X-axis indicates the average number of words in the summaries produced by each model. As outlined in Section \ref{sec:Metrics}, our goal is to identify the models that achieve the highest possible ROUGE score with the shortest summary length. Additionally, we establish a human baseline, calculated on the agreement between the human annotators detailed in Section \ref{sec:DataQuality}, and compare it against the gold summaries in the dataset. Extended results in numerical format are presented in Appendix \ref{sec:extendedResults}.

\subsection{Zero-Shot summarization}

First, we evaluate a wide range of Large Language Models (LLMs) spanning from 2 billion to 70 billion parameters on the NoticIA dataset in a zero-shot setting. This means that we use the models in their original form without any specific task fine-tuning. The models are prompted using the instructions described in Section \ref{sec:TaskFormualation}.

\input{Tablas/ErrorAnalysis}

The results indicate that Llama-3-70B-Instruct generates the highest quality summaries, with GPT-4o and GPT-4 following closely. Both Nous-Hermes-2-SOLAR-10.7B and OpenChat-3.5-0106, with 10.7 billion and 7 billion parameters respectively, deliver remarkably good performance despite being significantly smaller than the 70 billion parameter Llama-3 model. This makes these two models preferred the options when considering computational resource constraints.

The zero-shot results indicate that for this task, the number of model parameters holds little significance compared to the pre-training data. For instance, Llama-2-70b-chat-hf, a 70 billion parameter model, underperforms compared to the OpenChat-3.5-0106 model, which has only 7 billion parameters. The models that consistently achieve better performance share a common feature: they are trained on very large high-quality datasets. For example, the Nous-Hermes series of models have been trained using the OpenHermes dataset \cite{OpenHermes2.5}, which comprises 1 million instructions generated mainly with GPT-4. Similarly, OpenChat-3.5-0106 was also trained using data generated by GPT-4 and GPT-3.5 \cite{openchat}. Deepseek \cite{deepseek} was trained with 1.5 million instructions. Further analysis of the publicly available OpenHermes 2.5 dataset through the Lilac platform\footnote{\url{https://lilacai-lilac.hf.space/datasets\#lilac/OpenHermes-2.5}} reveals that this dataset contains at least 25,749 examples of summarization, a significant portion of which are article summaries. Therefore, this suggests that the best-performing models in our dataset are those that have been exposed to a large amount of summary examples during fine-tuning. It's important to note that the OpenHermes dataset predominantly contains instructions in English, with only 62 of the 1 million instructions being in Spanish, and these are specific examples of translations. This highlights the models' remarkable proficiency in transfer learning, demonstrating their ability to effectively apply knowledge learned from English data to tasks in Spanish.

\subsection{Error analysis}
To better understand the models' performance on our dataset, we conducted an error analysis to identify the circumstances and reasons why the models failed. We identified three different error sources for which we provide examples in Table \ref{tab:errorAnalysis}:

\paragraph{Producing a summary of the whole article instead of following the guidelines:} We identified that some models, instead of adhering to the provided guidelines and producing a concise summary based on the headline, simply summarize the entire article. Table \ref{tab:errorAnalysis} shows the summaries generated by Nous-Hermes-2-Yi-34B and Llama-2-70b-chat-hf along with the Gold summary. The former generates a summary that answers the question, while the latter produces an overly extensive summary of the whole article. We attribute this to the lower proficiency of some models in following instructions. As previously discussed, Nous-Hermes-2-Yi-34B, having been trained with 1 million instruction examples, has acquired much better proficiency at following instructions and adhering to the guidelines described in such instructions. The models in Figure \ref{fig:Results} which generate the longest summaries, all suffer from this issue.

\paragraph{Not concise enough summaries:} The best-performing models in our dataset, which follow the guidelines and produce summaries that address the clickbait headline, still generate summaries that are, on average, twice as long as those produced by humans. Even when explicitly prompted to produce an answer using the minimum number of words possible, and clarifying that the answer does not need to be a complete sentence, they still output a complete sentence. This likely reflects a bias from the instruction examples used during the models' fine-tuning phase. All models struggle to produce very short answers. Nevertheless, it is possible that further refining the prompt used in our experiments, or employing a multi-step inference process where the models are asked to reduce the number of words in their summaries, could mitigate this issue.

\paragraph{Failure to understand the headline or text:} This issue arises when the model does not correctly comprehend the clickbait headline or the information in the text. As illustrated in Table \ref{tab:errorAnalysis}, for the headline \textit{``¡Isabel Preysler estará en El Hormiguero! Estos son los próximos invitados''}, which asks about the upcoming guests on a Spanish TV program, OpenChat-3.5-0106, rather than listing the invited individuals, generates a summary that details the topic of discussion for a guest. This indicates a misunderstanding of the headline. Similarly, in response to the question \textit{``¿Qué pilotos de F1 acaban contrato en 2024?''},  which is accompanied by an article detailing the contract status of every Formula 1 driver, we found that none of the models could generate the correct list of drivers whose contracts end in 2024. In this case, this indicates a misunderstanding of the article body. Although the best-performing models show significantly fewer instances in which they fail to understand the headline or text, we found evidence of this issue across all models. In this case, the issue is related to the text comprehension capabilities of the models.

\subsubsection{Evaluation Consistency with Different Human Annotations}
\label{sec:Consistency}
\begin{figure}[tb]
   \centering
   \includegraphics[width=\linewidth]{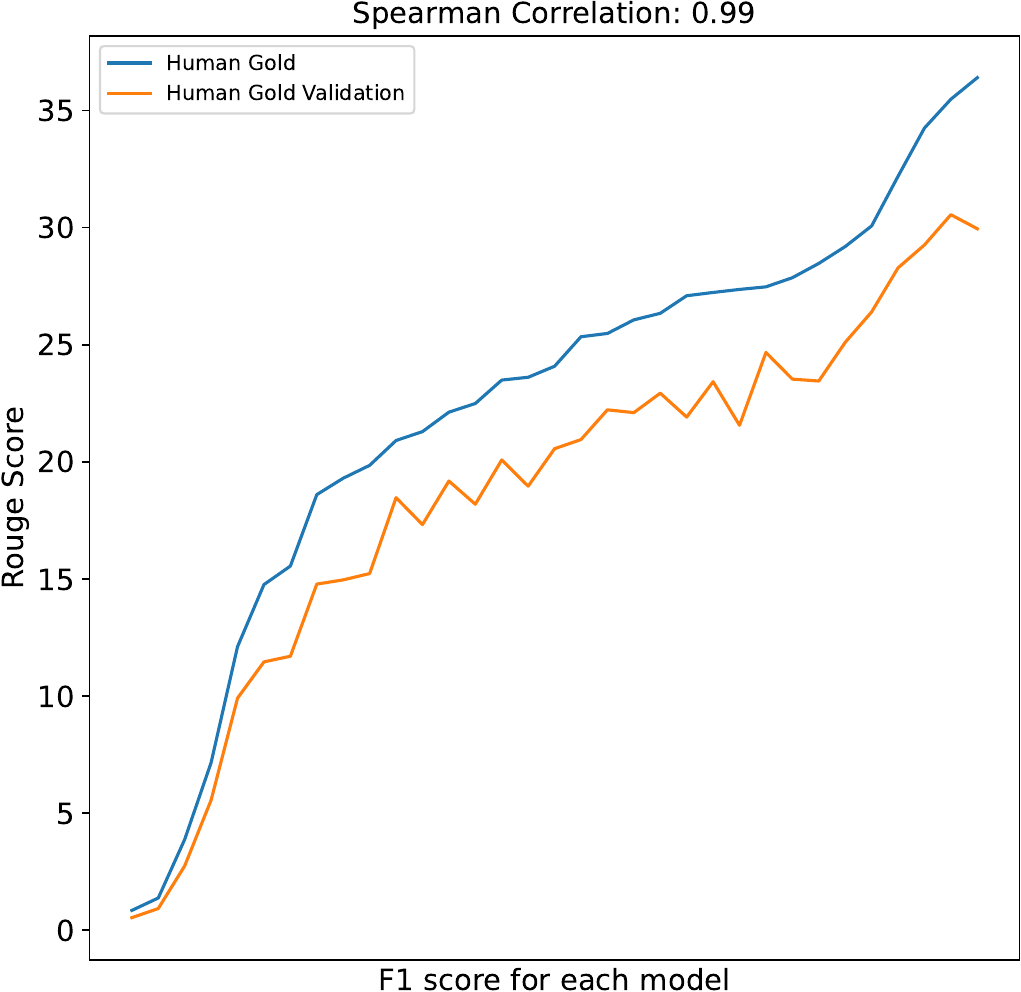}
   \caption{ROUGE scores of the models when evaluated against the gold summaries and the validation summaries produced by the second annotator.}
   \label{fig:modelsvshuman1and2}
 \end{figure}

To further assess the quality of our dataset and the reliability of the human annotations, we evaluated the models in zero-shot settings against both sets of human annotations described in Section \ref{sec:DataQuality}. The results are presented in Figure \ref{fig:modelsvshuman1and2}. The summaries from the second set of annotations were, on average, slightly shorter than those from the first set. This led to lower ROUGE scores for the models when compared with the second set of human annotations. Nevertheless, we observed a nearly perfect correlation in the scores (Spearman correlation of 0.99) when evaluated against both sets of human annotations. This indicates that the gold summaries in our dataset are robust and correlate well with those generated by different annotators following the same guidelines. It also demonstrates that humans can achieve a very high level of agreement on this task.

\subsection{\includegraphics[width=0.7cm]{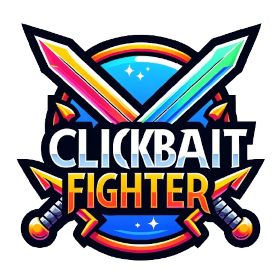} Clickbait Fighter a Task specific model}

In this effort, we fine-tune three task-specific models using the 700 training examples in the NoticIA dataset. We fine-tune three models of different sizes: ClickbaitFighter-10B, which is based on Nous-Hermes-2-SOLAR-10.7B, ClickbaitFighter-7B, which is based on openchat-3.5-0106, and ClickbaitFighter-2B, which is based on Gemma-2B-IT. We fine-tune all the parameters of the models using bfloat16 precision. More details about the fine-tuning process are available in Appendix \ref{sec:finetunedetails}. The results of the models are displayed in Figure \ref{fig:Results}. Despite being a small model, ClickbaitFighter-2B performs better than any evaluated LLM in the zero-shot setting, making this model ideal for situations where computational resources are constrained. On the other hand, ClickbaitFighter-10B and 7B achieve a summary quality close to the human baseline. Task-specific models overcome the struggle of producing concise summaries and manage to produce summaries almost twice as short as those from models in zero-shot settings.

\paragraph{How many training samples do we need?} 
\begin{figure}[tb]
   \centering
   \includegraphics[width=\linewidth]{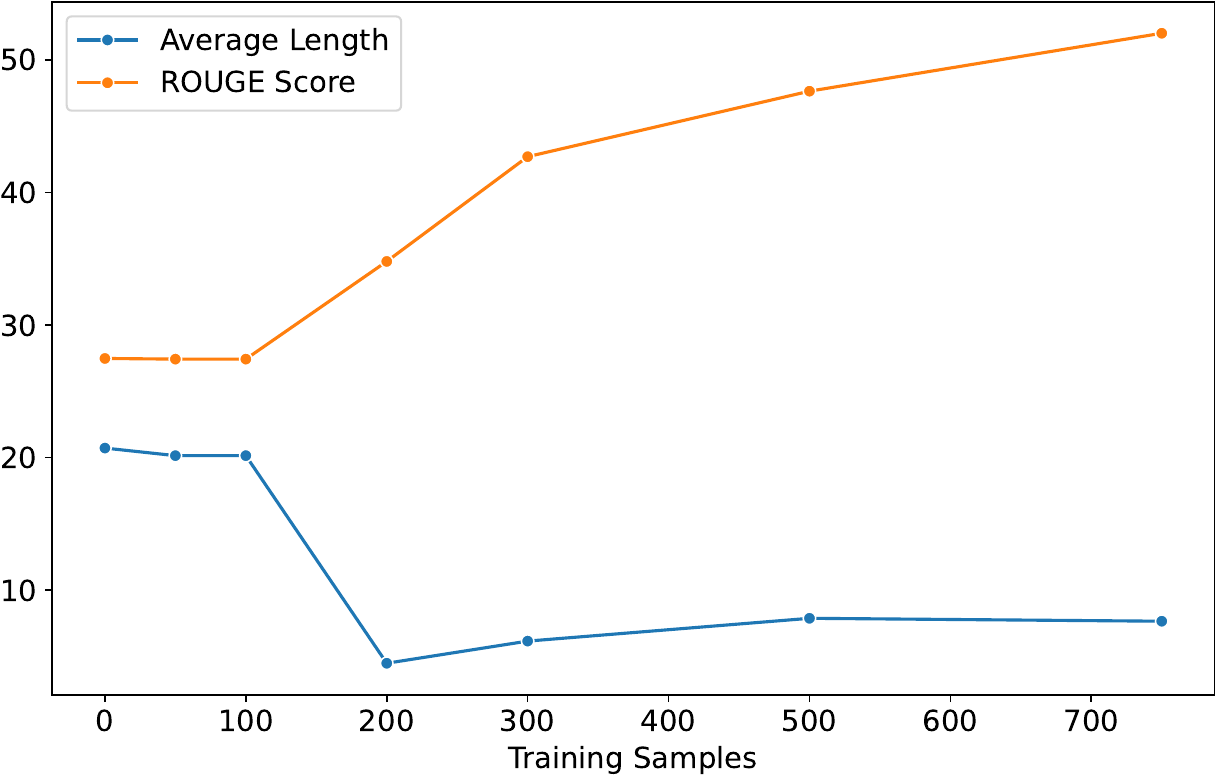}
   \caption{Comparison of Average Length and ROUGE Score when fine-tuning Nous-Hermes-2-SOLAR-10.7B with different amounts of training samples.}
   \label{fig:samplesvsperformance}
 \end{figure}

In Figure \ref{fig:samplesvsperformance}, we analyze the performance of Nous-Hermes-2-SOLAR-10.7B when trained with varying amounts of training samples. The results indicate that when trained with a small number of samples, the model can learn to produce concise summaries, reducing the average number of words in the output summary by 2.5. On the other hand, the ROUGE score continues to increase as more training samples are added and the model's performance does not show any signs of saturation. This suggests that extending the dataset beyond 750 training samples could further enhance model performance.

\section{Conclusions} \label{sec:conclusion}
In this work, we present \raisebox{-.25\height}{\includegraphics[width=0.5cm]{logo.png}}NoticIA, a dataset comprising 850 Spanish news articles featuring prominent clickbait headlines, each paired with high-quality, single-sentence summaries written by humans. To the best of our knowledge, NoticIA is the only clickbait dataset that provides headline-news body-ultrasummary triplets, which can be used to assess the performance of information retrieval systems aimed at extracting crucial information.

Our experiments demonstrate that NoticIA can effectively assess the text comprehension capabilities of large language models (LLMs) in the Spanish language. It also proves effective for training LLMs for the task of clickbait article summarization. Hence, our dataset holds the potential to contribute towards the development and evaluation of LLMs for the Spanish language.

In future work, we plan to further extend the dataset, as results indicate that expanding the dataset beyond 750 training samples could further enhance model performance.
\section*{Acknowledgements}

This work has been partially funded by:
\begin{itemize}
    \item DeepR3 (TED2021-130295B-C31) funded by MCIN/AEI/10.13039/501100011033 and European Union NextGeneration EU/PRTR.
    \item Disargue (TED2021-130810B-C21) MCIN/AEI/10.13039/501100011033 and European Union NextGenerationEU/PRTR.
    \item DeepKnowledge (PID2021-127777OB-C21) MCIN/AEI/10.13039/501100011033 and by FEDER, EU.
\end{itemize}

A demo of the ClickbaitFighter-7B model can be found in the SomosNLP Hugging Face organization. This demo was developed during the SomosNLP 2024 Hackathon\footnote{\url{https://somosnlp.org/}} \#Somos600M: \url{https://hf.co/spaces/somosnlp/NoticIA-demo}

\bibliographystyle{fullname}
\bibliography{EjemploARTsepln}

\appendix
\clearpage

\section{Fine tuning details}
\label{sec:finetunedetails}
We fine-tune all model parameters using bfloat16 precision. Utilizing DeepSpeed Zero 3 \cite{deepspeed}, we distribute the model, gradients, and optimizer across four A100 80GB GPUs. We employ the standard Next Token Prediction (NTP) loss for training our models. To prevent the loss associated with the article body tokens from overshadowing the loss of the summary output tokens, we compute the loss exclusively over the summary tokens. The hyperparameter settings are outlined in Table \ref{tab:hparams}. For all the experiments, we use greedy search to generate text.

\input{Tablas/hparams}

\section{Hardware used}
We conducted all our experiments on a machine equipped with four NVIDIA A100 GPUs, each with 80GB of memory, interconnected via NVLink. The machine features two AMD EPYC 7513 32-Core Processors and 1TB (1024GB) of RAM.

\section{NoticIA Dataset: Article sources} 
 \label{apx:Sources}

\begin{figure}[ht]
   \centering
   \includegraphics[width=\textwidth,height=0.94\textheight,keepaspectratio]{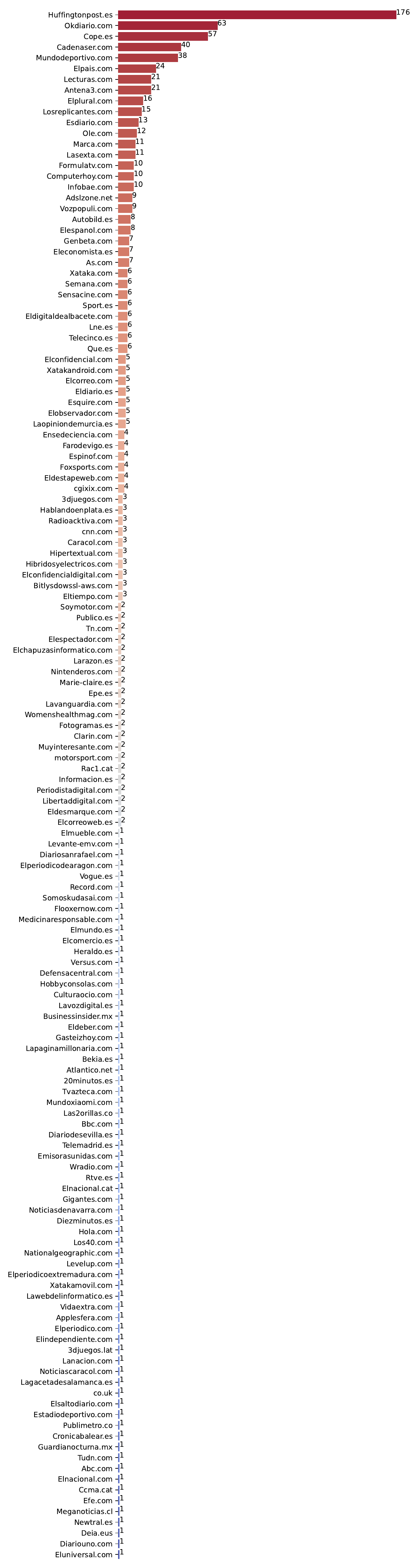}
   \caption{Sources from which the articles in the dataset have been gathered.}
   \label{fig:NoticiaSourceCount}
 \end{figure}

Figure \ref{fig:NoticiaSourceCount} lists all the sources from which we gathered the news articles. The articles were not evenly sampled from the news outlets; therefore, this table does not indicate the number of clickbait articles published by different newspapers. Instead, more popular news sources tend to have more appearances in our dataset.
 
 \clearpage

\section{Model details}
\label{apx:ModelDetails}

Table \ref{tab:model_sources} lists all the models we have evaluated, grouped by model families according to the base model on which they were trained. Each model has undergone instruction-following fine-tuning. We also provide the parameter count for each model, a citation for the model description, and the URL from which we downloaded them. All the models are open-source, except the ChatGPT model family. 

\input{Tablas/models}

\clearpage

\section{Extended Results}
\label{sec:extendedResults}
Table \ref{table:results} displays the results for the assessed models in numerical format. Additionally, we include models that, due to their significantly low performance, were not featured in the main results section.


 \section{NoticIA Dataset: Annotator Guidelines}
 \label{apx:guidelines}
\begin{figure*}[ht]
   \centering
   \includegraphics[width=\textwidth,height=0.94\textheight,keepaspectratio]{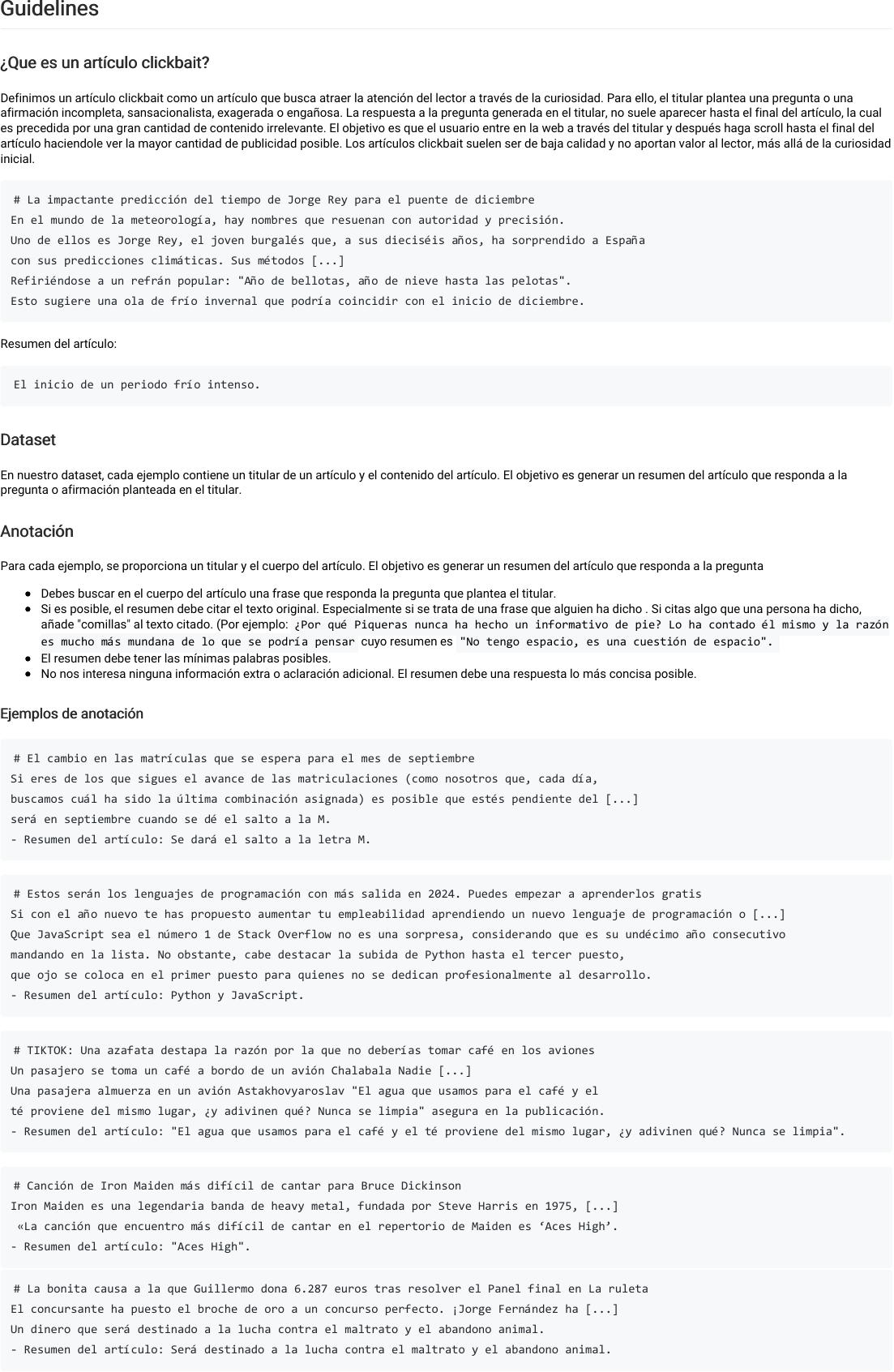}
   \caption{Guidelines provided to the annotators.}
   \label{fig:Guidelines}
 \end{figure*}

Figure \ref{fig:Guidelines} displays in detail the guidelines that were provided to the human annotators for generating summaries. These guidelines encompass a definition of ``clickbait", the specific instructions for annotation, and several examples. Initially, we asked the annotators to summarize a few examples. After inspecting these initial summaries, we revised and updated the guidelines to arrive at the final refined version. 
\newpage
\input{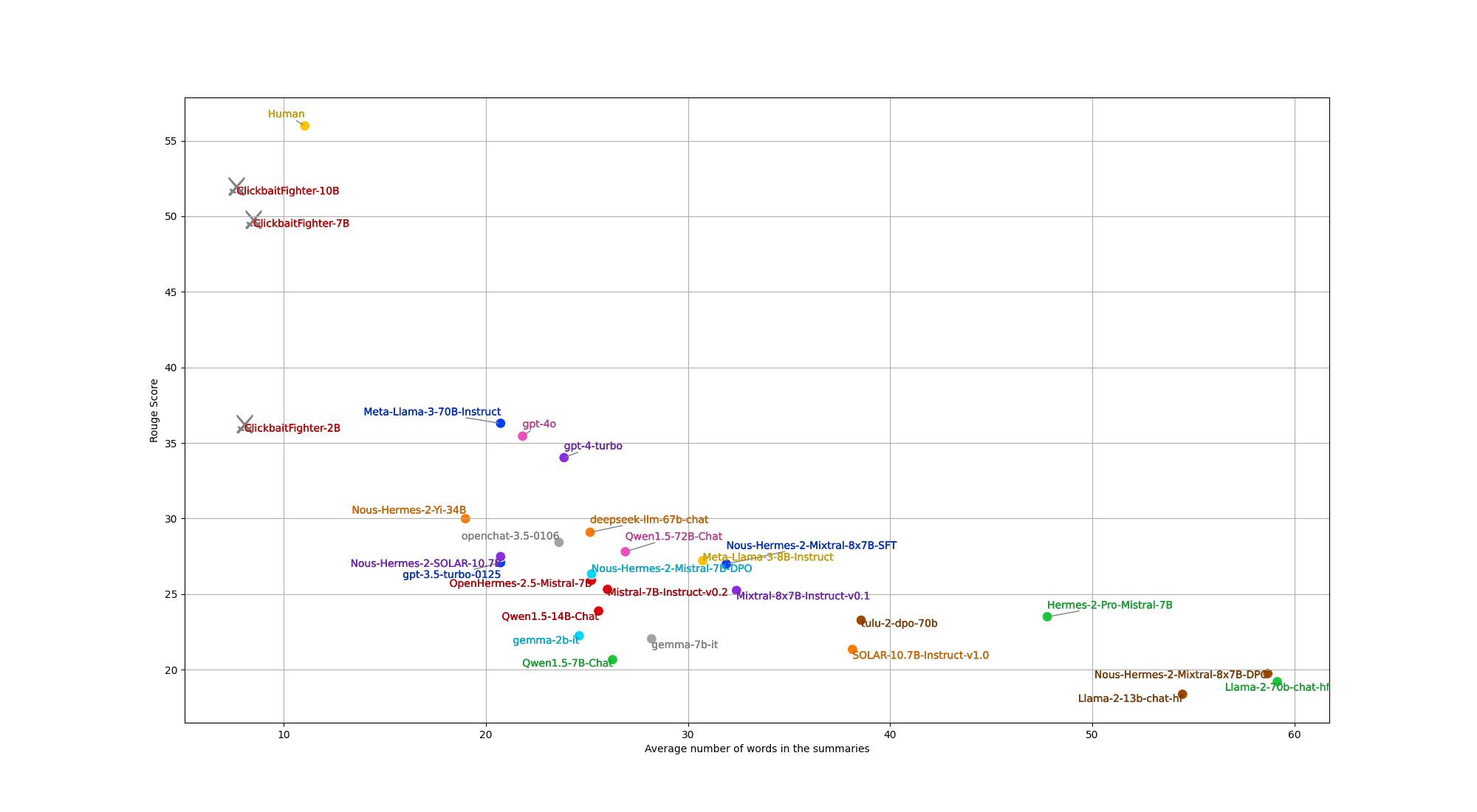}

\clearpage

\end{document}

%% file: Tablas/dataset_lens.tex
\begin{table}[t]
\centering
\adjustbox{max width=\linewidth}{
\begin{tabular}{@{}lllll@{}}
\toprule
                   & Train & Dev & Test & Total \\ \midrule
Article no.        & 700   & 50  & 100   & 850   \\
Avg. Headline Len. & 16    & 17  & 17   & 17    \\
Avg. Article Len.  & 544   & 663 & 549  & 552   \\
Avg. Summary Len.  & 12    & 11  & 11   & 12    \\ \bottomrule
\end{tabular}
}
\caption{Average word counts for the headlines, article bodies, and summaries.}
\label{tab:dataset_len}
\end{table}

%% file: Tablas/dataset_validation.tex
\begin{table}[t]
\centering
\adjustbox{max width=.5\textwidth}{
\begin{tabular}{@{}lcc@{}}
\toprule
                  & Test     & Validation     \\ \midrule
Rouge1            & \multicolumn{2}{c}{56.73} \\
Rouge2            & \multicolumn{2}{c}{43.05} \\
RougeL            & \multicolumn{2}{c}{55.59} \\ \midrule
Avg. Summary len. & 11       & 8              \\ \bottomrule
\end{tabular}
}
\caption{Validation stats between the summaries in the test split and the summaries produced by another independent annotator.}
\label{tab:dataset_validation}
\end{table}

%% file: Tablas/human_examples.tex
\begin{table*}[htb]
\centering
\adjustbox{max width=.9\textwidth}{
\begin{tabular}{@{}llp{8cm}@{}}
\toprule
                   & Annotator A & Annotator B  \\ \midrule
Same information        & Tenemos entre 30 y 45 minutos.   & Entre 30 y 45 minutos.     \\
More precise information & El regreso a clases para muchos estudiantes.    & El regreso a clases.      \\
Additional information  & Tener que saltar el potro.   & ``Cuando pasé a bachillerato, al BUP, estaba el potro y pasé unos días pensando que tenía saltar el potro y hacer deporte, muy malos".    \\
Shared information  & No le ha gustado por que ``es de vieja".    & ``El abrigo es de vieja", ha opinado la mujer.      \\ \bottomrule
\end{tabular}
}
\caption{Examples of partially overlapping answers by human annotators.}
\label{tab:human_examples}
\end{table*}

%% file: Tablas/models_small.tex
\begin{table*}[htb]
\centering
\adjustbox{max width=0.75\linewidth}{
\begin{tabular}{@{}llll@{}}
\toprule
Model Family              & Model name                     & Parameter Count & Citation                  \\ \midrule
Deepseek                  & deepseek-llm-67b-chat          & 67B             & \cite{deepseek}           \\\midrule
\multirow{2}{*}{Gemma}    & gemma-2b-it                    & 2.51B           & \cite{gemmateam2024gemma} \\
                          & gemma-7b-it                    & 8.54B           & \cite{gemmateam2024gemma} \\\midrule
\multirow{3}{*}{Llama2}   & Llama-2-70b-chat-hf            & 70B             & \cite{llama2}             \\
                          & Llama-2-13b-chat-hf            & 13B             & \cite{llama2}             \\ 
                          & tulu-2-dpo-70b                 & 70B             & \cite{tulu}               \\ \midrule
\multirow{2}{*}{Llama3}   & Meta-Llama-3-8B-Instruct            & 70B             & \cite{llama3modelcard}             \\
                          & Meta-Llama-3-8B-Instruct            & 8B             & \cite{llama3modelcard}             \\ \midrule
                          
\multirow{5}{*}{Mistral}  & Mistral-7B-Instruct-v0.2       & 7B              & \cite{mistral}            \\
                          & Nous-Hermes-2-Mistral-7B-DPO   & 7B              & \cite{OpenHermes2.5}      \\
                          & Nous-Hermes-2-Pro-Mistral-7B   & 7B              & \cite{OpenHermes2.5}      \\
                          & openchat-3.5-0106              & 7B              & \cite{openchat}           \\
                          & OpenHermes-2.5-Mistral-7B      & 7B              & \cite{OpenHermes2.5}      \\  \midrule
\multirow{2}{*}{Solar}    & SOLAR-10.7B-Instruct-v1.0      & 10.7B           & \cite{solar}              \\
                          & Nous-Hermes-2-SOLAR-10.7B      & 10.7B           & \cite{OpenHermes2.5}      \\\midrule
\multirow{3}{*}{Mixtral}  & Mixtral-8x7B-Instruct-v0.1     & 46.7B           & \cite{mixtral}            \\
                          & Nous-Hermes-2-Mixtral-8x7B-DPO & 46.7B           & \cite{OpenHermes2.5}      \\
                          & Nous-Hermes-2-Mixtral-8x7B-SFT & 46.7B           & \cite{OpenHermes2.5}      \\\midrule
\multirow{3}{*}{Yi}       & Nous-Hermes-2-Yi-34B           & 34B             & \cite{OpenHermes2.5}      \\
                          & Yi-6B-Chat                     & 6B              & \cite{ai2024yi}           \\
                          & Yi-34B-Chat                    & 34B             & \cite{ai2024yi}           \\\midrule
\multirow{4}{*}{QWEN}     & Qwen1.5-7B-Chat                & 7B              & \cite{qwen}               \\
                          & Qwen1.5-14B-Chat               & 14B             & \cite{qwen}               \\
                          & Qwen1.5-72B-Chat               & 72B             & \cite{qwen}               \\\midrule
\multirow{2}{*}{ChatGPT}  & gpt-4o-2024-05-13            & unknown         & \cite{gpt4o}               \\
                          & gpt-4-turbo-2024-04-09       & unknown         & \cite{gpt4}               \\
                          & gpt-3.5-turbo-0125           & unknown         & \cite{chatgpt}            \\ \bottomrule
\end{tabular}
}
\caption{Model that we use in our experiments. Model families group the models we use in our experiments, based on the base model on which they were trained.}
\label{tab:model_sources_small}
\end{table*}

%% file: Tablas/ErrorAnalysis.tex
\begin{table*}[htb]
    \centering
    \adjustbox{max width=.9\linewidth}{
    \begin{tabular}{@{}l|p{20cm}@{}}
    \toprule
    \multicolumn{2}{c}{Producing a summary of the whole article instead of following the guidelines} \\
    \midrule
    Headline & Los turistas británicos alucinan con el nuevo destino turístico español por su ron \\
    Gold & La isla de La Palma. \\
    Nous-Hermes-2-Yi-34B & Los turistas británicos se enamoran de La Palma por su microclima y gastronomía local. \\
    LLama-2-70b-chat-hf & Al rededor de 150.000 turistas llegan cada año, sólo una fracción de los millones de británicos que acuden en masa a sus hermanas mayores de Gran Canaria, Tenerife y Lanzarote\". Esta frase revela que La Palma es un destino turístico menos masivo que otras islas Canarias, lo que la hace atractiva para aquellos que buscan un lugar más tranquilo y auténtico. Además, se menciona la excelente gastronomía local, incluyendo el ron elaborado con caña de azúcar en la isla durante cientos de años, lo que sugiere que los visitantes pueden disfrutar de una experiencia culinaria única y auténtica. \\
    \midrule
    \multicolumn{2}{c}{Summaries not concise enough} \\ 
    \midrule
    Headline & La comedia familiar y navideña que ya está en Netflix y puedes ver en estas fiestas \\
    Gold & Mejor Navidad ¡imposible! \\
    Nous-Hermes-2-Yi-34B & La comedia familiar y navideña ``Mejor Navidad ¡Imposible!' ' ya está disponible en Netflix. \\
    GPT-4-0125-preview & ``Mejor Navidad ¡imposible! es la comedia familiar y navideña lanzada por Netflix el 16 de noviembre de 2023.' ' \\
    \midrule
    \multicolumn{2}{c}{Failure to understand the Headline or Text. } \\
    \midrule
    Headline & ¡Isabel Preysler estará en El Hormiguero! Estos son los próximos invitados \\
    Gold & Malú, Pedro Alonso, Begoña Vargas e Isabel Preysler. \\
    Nous-Hermes-2-Yi-34B & Isabel Preysler, Malú, Pedro Alonso y Begoña Vargas serán invitados en El Hormiguero la próxima semana. \\
    openchat-3.5-0106 & Isabel Preysler hablará sobre ``Isabel Preysler: mi Navidad' ' en Disney+. \\
    \midrule
    Headline & ¿Qué pilotos de F1 acaban contrato en 2024? \\
    Gold & Sergio Pérez, Carlos Sainz,  Fernando Alonso,  Esteban Ocon, Alexander Albon, Logan Sargeant, Yuki Tsunoda, Daniel Ricciardo,  Valtteri Bottas, Guanyu Zhou,  Kevin Magnussen y Nico Hulkenberg. \\
    Nous-Hermes-2-Yi-34B & Más de la mitad de la parrilla de F1 acaba contrato en 2024. \\
    GPT-4-0125-preview & Sergio Pérez, Carlos Sainz, Esteban Ocon, Alexander Albon, Logan Sargeant, Daniel Ricciardo, Yuki Tsunoda, Valtteri Bottas y Guanyu Zhou finalizan sus contratos en 2024. \\
    \bottomrule
    \end{tabular}}
    \caption{Error analysis, examples of the different errors produced by the models.}
    \label{tab:errorAnalysis}
\end{table*}

%% file: Tablas/hparams.tex
\begin{table}[htb]
\centering
\adjustbox{max width=\linewidth}{
\begin{tabular}{@{}lcc@{}}
\toprule
                & ClickbaitFighter-2B & ClickbaitFighter-7B/10B \\ \midrule
Batch Size      & 64                  & 64                   \\
Optimizer       & AdamW               & AdamW                \\
Scheduler       & Cosine              & Cosine               \\
Learning Rate   & 0.000005            & 0.000005             \\
Weight Decay    & 0.0                 & 0.0                  \\
Warmup Ratio    & 0.1                 & 0.1                  \\
adam\_beta1     & 0.9                 & 0.9                  \\
adam\_beta2     & 0.95                & 0.95                 \\
adam\_epsilon   & 1e-12               & 1e-12                \\
Epochs          & 5                   & 3                    \\
Sequence Length & 8192                & 8192                 \\
Compute Type           & Bfloat16            & Bfloat16             \\ \bottomrule
\end{tabular}
}
\caption{Detailed Hyperparameter Setting.}
\label{tab:hparams}
\end{table}

%% file: Tablas/models.tex
\begin{table}[hb]
\centering
\adjustbox{max width=\textwidth}{
\begin{tabular}{@{}lllll@{}}
\toprule
Model Family              & Model name                     & Parameter Count & Citation                  & URL                                                             \\ \midrule
Deepseek                  & deepseek-llm-67b-chat          & 67B             & \cite{deepseek}           & \url{https://hf.co/deepseek-ai/deepseek-llm-67b-chat}           \\
\multirow{2}{*}{Gemma}    & gemma-2b-it                    & 2.51B           & \cite{gemmateam2024gemma} & \url{https://hf.co/google/gemma-2b-it }                         \\
                          & gemma-7b-it                    & 8.54B           & \cite{gemmateam2024gemma} & \url{https://hf.co/google/gemma-7b-it}                          \\\midrule
\multirow{6}{*}{Llama2}   & Llama-2-70b-chat-hf            & 70B             & \cite{llama2}             & \url{https://hf.co/meta-llama/Llama-2-70b-chat-hf}              \\
                          & Llama-2-13b-chat-hf            & 13B             & \cite{llama2}             & \url{https://hf.co/meta-llama/Llama-2-13b-chat-hf}              \\
                          & Llama-2-7b-chat-hf             & 7B              & \cite{llama2}             & \url{https://hf.co/meta-llama/Llama-2-7b-chat-hf}               \\
                          & Nous-Hermes-2-Llama-2-70B.yaml & 70B             & \cite{OpenHermes2.5}      & \url{https://hf.co/NousResearch/Nous-Hermes-2-Llama-2-70B}      \\
                          & TinyLlama-1.1B-Chat-v1.0       & 1.1B            & \cite{tinyllama}          & \url{https://hf.co/TinyLlama/TinyLlama-1.1B-Chat-v1.0}          \\
                          & tulu-2-dpo-70b                 & 70B             & \cite{tulu}               & \url{https://hf.co/allenai/tulu-2-dpo-70b}                      \\\midrule
\multirow{2}{*}{Llama3}   & Meta-Llama-3-8B-Instruct       & 70B             & \cite{llama3modelcard}    & \url{https://hf.co/meta-llama/Meta-Llama-3-8B-Instruct}         \\
                          & Meta-Llama-3-8B-Instruct       & 8B             & \cite{llama3modelcard}     & \url{https://hf.co/meta-llama/Meta-Llama-3-70B-Instruct}        \\ \midrule
\multirow{5}{*}{Mistral}  & Mistral-7B-Instruct-v0.2       & 7B              & \cite{mistral}            & \url{https://hf.co/mistralai/Mistral-7B-Instruct-v0.2}          \\
                          & Nous-Hermes-2-Mistral-7B-DPO   & 7B              & \cite{OpenHermes2.5}      & \url{https://hf.co/NousResearch/Nous-Hermes-2-Mistral-7B-DPO}   \\
                          & Nous-Hermes-2-Pro-Mistral-7B   & 7B              & \cite{OpenHermes2.5}      & \url{https://hf.co/NousResearch/Hermes-2-Pro-Mistral-7B}        \\
                          & openchat-3.5-0106              & 7B              & \cite{openchat}           & \url{https://hf.co/openchat/openchat-3.5-0106}                  \\
                          & OpenHermes-2.5-Mistral-7B      & 7B              & \cite{OpenHermes2.5}      & \url{https://hf.co/teknium/OpenHermes-2.5-Mistral-7B}           \\
                          & zephyr-7b-beta                 & 7B              & \cite{Zephyr}             & \url{https://hf.co/HuggingFaceH4/zephyr-7b-beta}                \\\midrule
\multirow{2}{*}{Solar}    & SOLAR-10.7B-Instruct-v1.0      & 10.7B           & \cite{solar}              & \url{https://hf.co/upstage/SOLAR-10.7B-Instruct-v1.0}           \\
                          & Nous-Hermes-2-SOLAR-10.7B      & 10.7B           & \cite{OpenHermes2.5}      & \url{https://hf.co/NousResearch/Nous-Hermes-2-SOLAR-10.7B}      \\\midrule
\multirow{3}{*}{Mixtral}  & Mixtral-8x7B-Instruct-v0.1     & 46.7B           & \cite{mixtral}            & \url{https://hf.co/mistralai/Mixtral-8x7B-Instruct-v0.1}        \\
                          & Nous-Hermes-2-Mixtral-8x7B-DPO & 46.7B           & \cite{OpenHermes2.5}      & \url{https://hf.co/NousResearch/Nous-Hermes-2-Mixtral-8x7B-DPO} \\
                          & Nous-Hermes-2-Mixtral-8x7B-SFT & 46.7B           & \cite{OpenHermes2.5}      & \url{https://hf.co/NousResearch/Nous-Hermes-2-Mixtral-8x7B-SFT} \\\midrule
\multirow{3}{*}{StableLM} & Nous-Capybara-3B-V1.9          & 2.8B            & \cite{OpenHermes2.5}      & \url{https://hf.co/NousResearch/Nous-Capybara-3B-V1.9}          \\
                          & rocket-3B                      & 2.8B            & \cite{rocket-3B}          & \url{https://hf.co/pansophic/rocket-3B}                         \\
                          & stablelm-zephyr-3b             & 3B              & \cite{stablelm-zephyr}    & \url{https://hf.co/stabilityai/stablelm-zephyr-3b}              \\\midrule
\multirow{3}{*}{Yi}       & Nous-Hermes-2-Yi-34B           & 34B             & \cite{OpenHermes2.5}      & \url{https://hf.co/NousResearch/Nous-Hermes-2-Yi-34B}           \\
                          & Yi-6B-Chat                     & 6B              & \cite{ai2024yi}           & \url{https://hf.co/01-ai/Yi-6B-Chat}                            \\
                          & Yi-34B-Chat                    & 34B             & \cite{ai2024yi}           & \url{https://hf.co/01-ai/Yi-34B-Chat}                           \\\midrule
\multirow{4}{*}{QWEN}     & Qwen1.5-7B-Chat                & 7B              & \cite{qwen}               & \url{https://hf.co/Qwen/Qwen1.5-7B-Chat}                        \\
                          & Qwen1.5-14B-Chat               & 14B             & \cite{qwen}               & \url{https://hf.co/Qwen/Qwen1.5-14B-Chat}                       \\
                          & Qwen1.5-72B-Chat               & 72B             & \cite{qwen}               & \url{https://hf.co/Qwen/Qwen1.5-72B-Chat}                       \\
                          & Smaug-72B-v0.1                 & 72B             & \cite{pal2024smaug}       & \url{https://hf.co/abacusai/Smaug-72B-v0.1}                     \\\midrule
\multirow{2}{*}{ChatGPT}  & gpt-4o-2024-05-13              & unknown         & \cite{gpt4o}              & \url{https://openai.com/}                                       \\
                          & gpt-4-turbo-2024-04-09         & unknown         & \cite{gpt4}               & \url{https://openai.com/}                                       \\ 
                          & gpt-3.5-turbo-0125             & unknown         & \cite{chatgpt}            & \url{https://openai.com/}                                       \\ \bottomrule
\end{tabular}
}
\noindent\makebox[0.9\textwidth]{%
  \parbox{\textwidth}{%
    \captionof{table}{Models that we use in our experiments. Model families group the models used in our experiments, based on the base model on which they were trained.}
    \label{tab:model_sources}
    }
    }

\end{table}

%% file: Tablas/Results.tex
\begin{table}[H]
\centering
\begin{adjustbox}{width=\linewidth}
\begin{tabular}{lrr}
\toprule
                    Model Name &  ROUGE Score &  Average Summary Length  \\
\midrule
Human & 55.98 & 11.03 \\
ClickbaitFighter-10B & 52.01 & 7.66 \\
ClickbaitFighter-7B & 49.81 & 8.48 \\
Meta-Llama-3-70B-Instruct & 36.30 & 20.72 \\
ClickbaitFighter-2B & 36.26 & 8.04 \\
gpt-4o & 35.45 & 21.81 \\
gpt-4-turbo & 34.03 & 23.86 \\
gpt-4-0125-preview & 32.00 & 23.82 \\
Nous-Hermes-2-Yi-34B & 29.99 & 18.98 \\
deepseek-llm-67b-chat & 29.09 & 25.15 \\
openchat-3.5-0106 & 28.42 & 23.61 \\
Qwen1.5-72B-Chat & 27.80 & 26.89 \\
Nous-Hermes-2-SOLAR-10.7B & 27.48 & 20.72 \\
Meta-Llama-3-8B-Instruct & 27.21 & 30.72 \\
gpt-3.5-turbo-0125 & 27.08 & 20.72 \\
Nous-Hermes-2-Mixtral-8x7B-SFT & 26.99 & 31.89 \\
Nous-Hermes-2-Mistral-7B-DPO & 26.33 & 25.22 \\
OpenHermes-2.5-Mistral-7B & 25.91 & 25.21 \\
Mistral-7B-Instruct-v0.2 & 25.31 & 26.01 \\
Mixtral-8x7B-Instruct-v0.1 & 25.24 & 32.39 \\
Qwen1.5-14B-Chat & 23.88 & 25.57 \\
Hermes-2-Pro-Mistral-7B & 23.50 & 47.77 \\
tulu-2-dpo-70b & 23.27 & 38.56 \\
gemma-2b-it & 22.24 & 24.61 \\
gemma-7b-it & 22.04 & 28.19 \\
SOLAR-10.7B-Instruct-v1.0 & 21.34 & 38.13 \\
Qwen1.5-7B-Chat & 20.66 & 26.26 \\
Nous-Hermes-2-Mixtral-8x7B-DPO & 19.74 & 58.68 \\
Llama-2-70b-chat-hf & 19.20 & 59.15 \\
Llama-2-13b-chat-hf & 18.37 & 54.46 \\
zephyr-7b-beta & 15.64 & 105.75 \\
Llama-2-7b-chat-hf & 14.65 & 73.50 \\
stablelm-zephyr-3b & 12.02 & 86.90 \\
Nous-Hermes-2-Llama-2-70B & 7.22 & 154.55 \\
TinyLlama-1.1B-Chat-v1.0 & 3.84 & 191.15 \\
Smaug-72B-v0.1 & 1.79 & 82.74 \\
rocket-3B & 0.92 & 228.43 \\

\bottomrule
\end{tabular}
\end{adjustbox}
\caption{ROUGE score and average summary lengths for all the models evaluated in our dataset.}
\label{table:results}
\end{table}

%% file: CLICKBAIT.bbl
\begin{thebibliography}{}

\bibitem[\protect\citename{01.AI \bgroup et al.\egroup }2024]{ai2024yi}
01.AI, A.~Young, B.~Chen, C.~Li, C.~Huang, G.~Zhang, G.~Zhang, H.~Li, J.~Zhu, J.~Chen, J.~Chang, K.~Yu, P.~Liu, Q.~Liu, S.~Yue, S.~Yang, S.~Yang, T.~Yu, W.~Xie, W.~Huang, X.~Hu, X.~Ren, X.~Niu, P.~Nie, Y.~Xu, Y.~Liu, Y.~Wang, Y.~Cai, Z.~Gu, Z.~Liu, and Z.~Dai.
\newblock 2024.
\newblock {Yi: Open Foundation Models by 01.AI}.

\bibitem[\protect\citename{AI@Meta}2024]{llama3modelcard}
AI@Meta.
\newblock 2024.
\newblock Llama 3 model card.

\bibitem[\protect\citename{Bai \bgroup et al.\egroup }2023]{qwen}
Bai, J., S.~Bai, Y.~Chu, Z.~Cui, K.~Dang, X.~Deng, Y.~Fan, W.~Ge, Y.~Han, F.~Huang, B.~Hui, L.~Ji, M.~Li, J.~Lin, R.~Lin, D.~Liu, G.~Liu, C.~Lu, K.~Lu, J.~Ma, R.~Men, X.~Ren, X.~Ren, C.~Tan, S.~Tan, J.~Tu, P.~Wang, S.~Wang, W.~Wang, S.~Wu, B.~Xu, J.~Xu, A.~Yang, H.~Yang, J.~Yang, S.~Yang, Y.~Yao, B.~Yu, H.~Yuan, Z.~Yuan, J.~Zhang, X.~Zhang, Y.~Zhang, Z.~Zhang, C.~Zhou, J.~Zhou, X.~Zhou, and T.~Zhu.
\newblock 2023.
\newblock {Qwen Technical Report}.
\newblock {\em CoRR}, abs/2309.16609.

\bibitem[\protect\citename{Bi \bgroup et al.\egroup }2024]{deepseek}
Bi, X., D.~Chen, G.~Chen, S.~Chen, D.~Dai, C.~Deng, H.~Ding, K.~Dong, Q.~Du, Z.~Fu, H.~Gao, K.~Gao, W.~Gao, R.~Ge, K.~Guan, D.~Guo, J.~Guo, G.~Hao, Z.~Hao, Y.~He, W.~Hu, P.~Huang, E.~Li, G.~Li, J.~Li, Y.~Li, Y.~K. Li, W.~Liang, F.~Lin, A.~X. Liu, B.~Liu, W.~Liu, X.~Liu, X.~Liu, Y.~Liu, H.~Lu, S.~Lu, F.~Luo, S.~Ma, X.~Nie, T.~Pei, Y.~Piao, J.~Qiu, H.~Qu, T.~Ren, Z.~Ren, C.~Ruan, Z.~Sha, Z.~Shao, J.~Song, X.~Su, J.~Sun, Y.~Sun, M.~Tang, B.~Wang, P.~Wang, S.~Wang, Y.~Wang, Y.~Wang, T.~Wu, Y.~Wu, X.~Xie, Z.~Xie, Z.~Xie, Y.~Xiong, H.~Xu, R.~X. Xu, Y.~Xu, D.~Yang, Y.~You, S.~Yu, X.~Yu, B.~Zhang, H.~Zhang, L.~Zhang, L.~Zhang, M.~Zhang, M.~Zhang, W.~Zhang, Y.~Zhang, C.~Zhao, Y.~Zhao, S.~Zhou, S.~Zhou, Q.~Zhu, and Y.~Zou.
\newblock 2024.
\newblock {DeepSeek {LLM:} Scaling Open-Source Language Models with Longtermism}.
\newblock {\em CoRR}, abs/2401.02954.

\bibitem[\protect\citename{Fr{\"o}be \bgroup et al.\egroup }2023]{froebe:2023d}
Fr{\"o}be, M., B.~Stein, T.~Gollub, M.~Hagen, and M.~Potthast.
\newblock 2023.
\newblock {{{S}em{E}val-2023 Task 5: Clickbait Spoiling}}.
\newblock In A.~K. Ojha, A.~S. Do{\u{g}}ru{\"o}z, G.~Da~San~Martino, H.~Tayyar~Madabushi, R.~Kumar, and E.~Sartori, editors, {\em Proceedings of the 17th International Workshop on Semantic Evaluation (SemEval-2023)}, pages 2275--2286, Toronto, Canada, July. Association for Computational Linguistics.

\bibitem[\protect\citename{Gemma-Team \bgroup et al.\egroup }2024]{gemmateam2024gemma}
Gemma-Team, T.~Mesnard, C.~Hardin, R.~Dadashi, S.~Bhupatiraju, S.~Pathak, L.~Sifre, M.~Rivière, M.~S. Kale, J.~Love, P.~Tafti, L.~Hussenot, A.~Chowdhery, A.~Roberts, A.~Barua, A.~Botev, A.~Castro-Ros, A.~Slone, A.~Héliou, A.~Tacchetti, A.~Bulanova, A.~Paterson, B.~Tsai, B.~Shahriari, C.~L. Lan, C.~A. Choquette-Choo, C.~Crepy, D.~Cer, D.~Ippolito, D.~Reid, E.~Buchatskaya, E.~Ni, E.~Noland, G.~Yan, G.~Tucker, G.-C. Muraru, G.~Rozhdestvenskiy, H.~Michalewski, I.~Tenney, I.~Grishchenko, J.~Austin, J.~Keeling, J.~Labanowski, J.-B. Lespiau, J.~Stanway, J.~Brennan, J.~Chen, J.~Ferret, J.~Chiu, J.~Mao-Jones, K.~Lee, K.~Yu, K.~Millican, L.~L. Sjoesund, L.~Lee, L.~Dixon, M.~Reid, M.~Mikuła, M.~Wirth, M.~Sharman, N.~Chinaev, N.~Thain, O.~Bachem, O.~Chang, O.~Wahltinez, P.~Bailey, P.~Michel, P.~Yotov, P.~G. Sessa, R.~Chaabouni, R.~Comanescu, R.~Jana, R.~Anil, R.~McIlroy, R.~Liu, R.~Mullins, S.~L. Smith, S.~Borgeaud, S.~Girgin, S.~Douglas, S.~Pandya, S.~Shakeri, S.~De, T.~Klimenko, T.~Hennigan, V.~Feinberg,
  W.~Stokowiec, Y.~hui Chen, Z.~Ahmed, Z.~Gong, T.~Warkentin, L.~Peran, M.~Giang, C.~Farabet, O.~Vinyals, J.~Dean, K.~Kavukcuoglu, D.~Hassabis, Z.~Ghahramani, D.~Eck, J.~Barral, F.~Pereira, E.~Collins, A.~Joulin, N.~Fiedel, E.~Senter, A.~Andreev, and K.~Kenealy.
\newblock 2024.
\newblock {Gemma: Open Models Based on Gemini Research and Technology}.

\bibitem[\protect\citename{Heiervang}2022]{Heiervang2022}
Heiervang, M.
\newblock 2022.
\newblock Abstractive title answering for clickbait content.
\newblock Master's thesis, University of Oslo.

\bibitem[\protect\citename{Intan~Maharani, Purwarianti, and Aji}2023]{Maharani2023}
Intan~Maharani, N.~P., A.~Purwarianti, and A.~F. Aji.
\newblock 2023.
\newblock {Low-Resource Clickbait Spoiling for Indonesian via Question Answering}.
\newblock In {\em 2023 10th International Conference on Advanced Informatics: Concept, Theory and Application (ICAICTA)}, pages 1--6.

\bibitem[\protect\citename{Ivison \bgroup et al.\egroup }2023]{tulu}
Ivison, H., Y.~Wang, V.~Pyatkin, N.~Lambert, M.~Peters, P.~Dasigi, J.~Jang, D.~Wadden, N.~A. Smith, I.~Beltagy, and H.~Hajishirzi.
\newblock 2023.
\newblock {Camels in a Changing Climate: Enhancing {LM} Adaptation with Tulu 2}.
\newblock {\em CoRR}, abs/2311.10702.

\bibitem[\protect\citename{Jiang \bgroup et al.\egroup }2023]{mistral}
Jiang, A.~Q., A.~Sablayrolles, A.~Mensch, C.~Bamford, D.~S. Chaplot, D.~de~Las~Casas, F.~Bressand, G.~Lengyel, G.~Lample, L.~Saulnier, L.~R. Lavaud, M.~Lachaux, P.~Stock, T.~L. Scao, T.~Lavril, T.~Wang, T.~Lacroix, and W.~E. Sayed.
\newblock 2023.
\newblock {Mistral 7B}.
\newblock {\em CoRR}, abs/2310.06825.

\bibitem[\protect\citename{Jiang \bgroup et al.\egroup }2024]{mixtral}
Jiang, A.~Q., A.~Sablayrolles, A.~Roux, A.~Mensch, B.~Savary, C.~Bamford, D.~S. Chaplot, D.~de~Las~Casas, E.~B. Hanna, F.~Bressand, G.~Lengyel, G.~Bour, G.~Lample, L.~R. Lavaud, L.~Saulnier, M.~Lachaux, P.~Stock, S.~Subramanian, S.~Yang, S.~Antoniak, T.~L. Scao, T.~Gervet, T.~Lavril, T.~Wang, T.~Lacroix, and W.~E. Sayed.
\newblock 2024.
\newblock {Mixtral of Experts}.
\newblock {\em CoRR}, abs/2401.04088.

\bibitem[\protect\citename{Kim \bgroup et al.\egroup }2023]{solar}
Kim, D., C.~Park, S.~Kim, W.~Lee, W.~Song, Y.~Kim, H.~Kim, Y.~Kim, H.~Lee, J.~Kim, C.~Ahn, S.~Yang, S.~Lee, H.~Park, G.~Gim, M.~Cha, H.~Lee, and S.~Kim.
\newblock 2023.
\newblock {{SOLAR} 10.7B: Scaling Large Language Models with Simple yet Effective Depth Up-Scaling}.
\newblock {\em CoRR}, abs/2312.15166.

\bibitem[\protect\citename{Kurenkov \bgroup et al.\egroup }2022]{johnson2022saved}
Kurenkov, A., T.~Mentor, Y.~Zhang, and O.~C. Johnson.
\newblock 2022.
\newblock {Saved You A Click: Automatically Answering Clickbait Titles}.
\newblock {\em ArXiv}, abs/2212.08196.

\bibitem[\protect\citename{Lin}2004]{lin-2004-ROUGE}
Lin, C.-Y.
\newblock 2004.
\newblock {ROUGE: A Package for Automatic Evaluation of Summaries}.
\newblock In {\em Text Summarization Branches Out}, pages 74--81, Barcelona, Spain, July. Association for Computational Linguistics.

\bibitem[\protect\citename{Liu \bgroup et al.\egroup }2021]{Liu2021}
Liu, T., K.~Yu, L.~Wang, X.~Zhang, and X.~Wu.
\newblock 2021.
\newblock {WCD: A New Chinese Online Social Media Dataset for Clickbait Analysis and Detection}.
\newblock In {\em 2021 7th IEEE International Conference on Network Intelligence and Digital Content (IC-NIDC)}, pages 368--372.

\bibitem[\protect\citename{Min \bgroup et al.\egroup }2023]{min-etal-2023-factscore}
Min, S., K.~Krishna, X.~Lyu, M.~Lewis, W.-t. Yih, P.~Koh, M.~Iyyer, L.~Zettlemoyer, and H.~Hajishirzi.
\newblock 2023.
\newblock {FA}ct{S}core: Fine-grained atomic evaluation of factual precision in long form text generation.
\newblock In H.~Bouamor, J.~Pino, and K.~Bali, editors, {\em Proceedings of the 2023 Conference on Empirical Methods in Natural Language Processing}, pages 12076--12100, Singapore, December. Association for Computational Linguistics.

\bibitem[\protect\citename{OpenAI}2023]{gpt4}
OpenAI.
\newblock 2023.
\newblock {GPT-4 Technical Report}.
\newblock {\em CoRR}, abs/2303.08774.

\bibitem[\protect\citename{OpenAI}2024a]{chatgpt}
OpenAI.
\newblock 2024a.
\newblock gpt-3.5-turbo-0125.

\bibitem[\protect\citename{OpenAI}2024b]{gpt4o}
OpenAI.
\newblock 2024b.
\newblock {Hello GPT-4o}.

\bibitem[\protect\citename{Pal \bgroup et al.\egroup }2024]{pal2024smaug}
Pal, A., D.~Karkhanis, S.~Dooley, M.~Roberts, S.~Naidu, and C.~White.
\newblock 2024.
\newblock {Smaug: Fixing Failure Modes of Preference Optimisation with DPO-Positive}.

\bibitem[\protect\citename{pansophic}2023]{rocket-3B}
pansophic.
\newblock 2023.
\newblock {A 3B parameter GPT-like model fine-tuned on a mix of publicly available datasets using DPO}.

\bibitem[\protect\citename{Potthast \bgroup et al.\egroup }2018]{potthast2018}
Potthast, M., T.~Gollub, K.~Komlossy, S.~Schuster, M.~Wiegmann, E.~P. Garces~Fernandez, M.~Hagen, and B.~Stein.
\newblock 2018.
\newblock {Crowdsourcing a Large Corpus of Clickbait on {T}witter}.
\newblock In E.~M. Bender, L.~Derczynski, and P.~Isabelle, editors, {\em Proceedings of the 27th International Conference on Computational Linguistics}, pages 1498--1507, Santa Fe, New Mexico, USA, August. Association for Computational Linguistics.

\bibitem[\protect\citename{Pujahari and Sisodia}2021]{pujahari2021}
Pujahari, A. and D.~S. Sisodia.
\newblock 2021.
\newblock {Clickbait detection using multiple categorisation techniques}.
\newblock {\em Journal of Information Science}, 47(1):118--128.

\bibitem[\protect\citename{Ren \bgroup et al.\egroup }2021]{deepspeed}
Ren, J., S.~Rajbhandari, R.~Y. Aminabadi, O.~Ruwase, S.~Yang, M.~Zhang, D.~Li, and Y.~He.
\newblock 2021.
\newblock {ZeRO-Offload: Democratizing Billion-Scale Model Training}.
\newblock In I.~Calciu and G.~Kuenning, editors, {\em 2021 {USENIX} Annual Technical Conference, {USENIX} {ATC} 2021, July 14-16, 2021}, pages 551--564. {USENIX} Association.

\bibitem[\protect\citename{Sepúlveda-Torres, Bonet-Jover, and Saquete}2023]{Sepulveda2023}
Sepúlveda-Torres, R., A.~Bonet-Jover, and E.~Saquete.
\newblock 2023.
\newblock {Detecting Misleading Headlines Through the Automatic Recognition of Contradiction in Spanish}.
\newblock {\em IEEE Access}, 11:72007--72026.

\bibitem[\protect\citename{stability.ai}2023]{stablelm-zephyr}
stability.ai.
\newblock 2023.
\newblock {Introducing Stable LM Zephyr 3B: A New Addition to Stable LM, Bringing Powerful LLM Assistants to Edge Devices}.

\bibitem[\protect\citename{Teknium}2023]{OpenHermes2.5}
Teknium.
\newblock 2023.
\newblock {OpenHermes 2.5: An Open Dataset of Synthetic Data for Generalist LLM Assistants}.

\bibitem[\protect\citename{Touvron \bgroup et al.\egroup }2023]{llama2}
Touvron, H., L.~Martin, K.~Stone, P.~Albert, A.~Almahairi, Y.~Babaei, N.~Bashlykov, S.~Batra, P.~Bhargava, S.~Bhosale, D.~Bikel, L.~Blecher, C.~Canton{-}Ferrer, M.~Chen, G.~Cucurull, D.~Esiobu, J.~Fernandes, J.~Fu, W.~Fu, B.~Fuller, C.~Gao, V.~Goswami, N.~Goyal, A.~Hartshorn, S.~Hosseini, R.~Hou, H.~Inan, M.~Kardas, V.~Kerkez, M.~Khabsa, I.~Kloumann, A.~Korenev, P.~S. Koura, M.~Lachaux, T.~Lavril, J.~Lee, D.~Liskovich, Y.~Lu, Y.~Mao, X.~Martinet, T.~Mihaylov, P.~Mishra, I.~Molybog, Y.~Nie, A.~Poulton, J.~Reizenstein, R.~Rungta, K.~Saladi, A.~Schelten, R.~Silva, E.~M. Smith, R.~Subramanian, X.~E. Tan, B.~Tang, R.~Taylor, A.~Williams, J.~X. Kuan, P.~Xu, Z.~Yan, I.~Zarov, Y.~Zhang, A.~Fan, M.~Kambadur, S.~Narang, A.~Rodriguez, R.~Stojnic, S.~Edunov, and T.~Scialom.
\newblock 2023.
\newblock {Llama 2: Open Foundation and Fine-Tuned Chat Models}.
\newblock {\em CoRR}, abs/2307.09288.

\bibitem[\protect\citename{Tunstall \bgroup et al.\egroup }2023]{Zephyr}
Tunstall, L., E.~Beeching, N.~Lambert, N.~Rajani, K.~Rasul, Y.~Belkada, S.~Huang, L.~von Werra, C.~Fourrier, N.~Habib, N.~Sarrazin, O.~Sanseviero, A.~M. Rush, and T.~Wolf.
\newblock 2023.
\newblock {Zephyr: Direct Distillation of {LM} Alignment}.
\newblock {\em CoRR}, abs/2310.16944.

\bibitem[\protect\citename{Wang \bgroup et al.\egroup }2023]{openchat}
Wang, G., S.~Cheng, X.~Zhan, X.~Li, S.~Song, and Y.~Liu.
\newblock 2023.
\newblock {OpenChat: Advancing Open-source Language Models with Mixed-Quality Data}.
\newblock {\em CoRR}, abs/2309.11235.

\bibitem[\protect\citename{Wang, Maslim, and Liu}2023]{Wang2023}
Wang, H.-C., M.~Maslim, and H.-Y. Liu.
\newblock 2023.
\newblock {CA-CD: context-aware clickbait detection using new Chinese clickbait dataset with transfer learning method}.
\newblock {\em Data Technologies and Applications}, 58(2), Jan.

\bibitem[\protect\citename{Zhang \bgroup et al.\egroup }2024]{tinyllama}
Zhang, P., G.~Zeng, T.~Wang, and W.~Lu.
\newblock 2024.
\newblock {TinyLlama: An Open-Source Small Language Model}.
\newblock {\em CoRR}, abs/2401.02385.

\bibitem[\protect\citename{Zheng, Yu, and Wu}2021]{ZHENG2021}
Zheng, J., K.~Yu, and X.~Wu.
\newblock 2021.
\newblock {A deep model based on Lure and Similarity for Adaptive Clickbait Detection}.
\newblock {\em Knowledge-Based Systems}, 214:106714.

\end{thebibliography}
